\documentclass[letterpaper, 10 pt, conference]{ieeeconf}  

\IEEEoverridecommandlockouts                              

\usepackage{soul}
\usepackage{xcolor}
\usepackage{hyperref}
\usepackage{cleveref}
\usepackage{graphicx}

\usepackage{flushend}

\title{\LARGE \bf Leveraging YOLO-World and GPT-4V LMMs for\\Zero-Shot Person Detection and Action Recognition in Drone Imagery 
}

\author{Christian Limberg$^{1}$, Artur Gon\c{c}alves$^{1}$, Bastien Rigault$^{1}$ and Helmut Prendinger$^{1}$
\thanks{*This work was supported by a fellowship within the IFI program of the German Academic Exchange Service (DAAD).}
\thanks{$^{1}$National Institute of Informatics (NII), Tokyo, Japan
        {\tt\small cnlimberg@gmail.com}}%
}

\begin{document}

\maketitle
\thispagestyle{empty}
\pagestyle{empty}

\begin{abstract}

In this article, we explore the potential of zero-shot Large Multimodal Models (LMMs) in the domain of drone perception. We focus on person detection and action recognition tasks and evaluate two prominent LMMs, namely YOLO-World and GPT-4V(ision) using a publicly available dataset captured from aerial views. 
Traditional deep learning approaches rely heavily on large and high-quality training datasets. However, in certain robotic settings, acquiring such datasets can be resource-intensive or impractical within a reasonable timeframe. The flexibility of prompt-based Large Multimodal Models (LMMs) and their exceptional generalization capabilities have the potential to revolutionize robotics applications in these scenarios.
Our findings suggest that YOLO-World demonstrates good detection performance. GPT-4V struggles with accurately classifying action classes but delivers promising results in filtering out unwanted region proposals and in providing a general description of the scenery. This research represents an initial step in leveraging LMMs for drone perception and establishes a foundation for future investigations in this area.

\end{abstract}

\section{INTRODUCTION}

Recent advances in Large Language Models (LLMs) have transformed many aspects of Machine Learning and AI~\cite{gpt3,gpt4}. Previously, the most common approach involved gathering datasets that captured small contexts within specific task domains. However, with the advent of foundation models such as LLMs, trained on much larger datasets, this paradigm has shifted. These models can now be utilized by providing them with prompts that specify the domain and task. Thanks to their strong generalization abilities, these foundation models can often be applied in a zero-shot manner~\cite{yang2023dawn}.
While LLMs were originally designed for processing text in Natural Language Processing (NLP) tasks, Large Multimodal Models (LMMs) have expanded their capabilities by incorporating additional modalities~\cite{2023gpt4v, copet2023simple, blattmann2023align} such as images, sounds, and videos.

In this article, we delve into the application of two recent image-based LMMs within a drone setting. Firstly, we examine the YOLO-World model~\cite{cheng2024yoloworld}, which facilitates prompt-based object detection. Secondly, we utilize the more general vision model GPT-4V~\cite{2023gpt4v} for classifying the detected region proposals.

An important challenge in aerial robotics is ensuring that drones operate reliably across a wide spectrum of potential failures. This necessitates the acquisition of a high-quality, problem-specific dataset, which can be resource-intensive or even impractical to obtain. Moreover, conventionally trained models tend to excel only within the confines of their training data. Minor variations in the environment, such as changes in weather, seasonal fluctuations, or geographical differences, can lead to a significant decline in the robot's reliability.

LMMs, trained on a broader contextual scope, may not achieve competitive performance compared to their traditional counterparts within their narrowly defined training contexts~\cite{gpt3}. However, their ability to generalize across domains, facilitated by their training on significantly broader ranges of data, enables them to better handle challenging conditions.

This preliminary study investigates the feasibility of applying YOLO-World and GPT-4V in a practical aerial robotic scenario involving person detection and action recognition. A real-world application could entail locating individuals in need following a disaster~\cite{daud2022applications, zhao2023agent}. Given the unpredictable nature of potential disasters, it is crucial to utilize a model with extensive generalization capabilities, capable of operating effectively across diverse settings.
Both YOLO-World and GPT-4V are zero-shot approaches prompted with text. This means they could potentially be deployed in unforeseen scenarios, as the text prompts can be adjusted quickly, enabling the equipped robot to adapt to entirely different objectives instantly.

This manuscript is structured as follows:
\Cref{relatedwork} provides an overview of the most important related works.
\Cref{okutama} discusses the publicly available Okutama-Action dataset that we are utilizing for our evaluation.
In \Cref{persondetection}, we focus on detecting persons using YOLO-World.
In \Cref{actionrecognition}, we apply GPT-4V on the detected region proposals to recognize the persons' actions.
Finally, \Cref{conclusion} summarizes our findings and concludes the paper.

\section{RELATED WORK}
\label{relatedwork}
From the very beginnings of computer vision research, object detection has been a prominent task of interest. First, hand-crafted features~\cite{2005histograms} were utilized for detecting and recognizing objects. With the rise of deep learning, convolutional neural networks, which derive features automatically from the training data, quickly overtook established handcrafted methods in terms of accuracy and robustness~\cite{krizhevsky2012imagenet}. ``Two-stage'' methods like R-CNN~\cite{2014rcnn} and R-FCN~\cite{2016rfcn} first detect candidate regions proposals, and then classify them.

Later, ``one-stage'' methods such as SSD~\cite{2016ssd} and YOLO~\cite{2016yolov1} established themselves, achieving higher processing speeds by detecting and classifying objects with one forward pass through the network. YOLO in particular has become extremely popular object detection method, with numerous iterations and variants~\cite{2018yolov3, bochkovskiy2020yolov4, 2023ultralyticsyolo, 2021yolox, wang2024yolov9}. New variants like YOLOX~\cite{2021yolox} and FCOS~\cite{2019fcos} have found improvements by moving away from the concept of fixed anchor boxes~\cite{limberg2022yolo}, integrating new data augmentation techniques and optimizing for new training objectives.

Still, the basic training paradigm of those models is the same compared with traditional handcrafted approaches. There is a fixed task definition and the need to acquire a data set that captures this task definition very accurately. Then the model is trained on a large part of the data and evaluated on a smaller fraction.

With the introduction of LLMs, there was also a shift in this training paradigm. OpenAI demonstrated this by autoregressively pre-training a large transformer-based model on a vast collection of internet text corpora~\cite{gpt3}, showcasing that the contained general knowledge can be effectively compressed. With a fine-tuning step the model can be tuned for a specific task. ChatGPT~\cite{RAY2023121} and GPT-4~\cite{gpt4} are trained in this post-training step with supervised data and with reinforcement learning from human feedback to realize a conversational agent that can be prompted with a wide range of tasks.

Recently, OpenAI has released the vision-enabled variant GPT-4V~\cite{2023gpt4v}, which shows promising capabilities on real-world visual understanding~\cite{yang2023dawn}. However, GPT-4V still underperforms in detecting humans within a drone context where humans are usually captured in a few pixels and recorded from steep angles. We could not replicate the person detection results shown in ~\cite{yang2023dawn} (pp.~42, Figure~27) for those difficult conditions.

To compensate for these shortcomings we decided to explore whether YOLO-World~\cite{cheng2024yoloworld} can be utilized for the detection task within this challenging application. YOLO-World is a very new approach within the YOLO family, published only two and a half weeks from the writing of this manuscript. While the approach incorporates traditional YOLO elements like the YOLO backbone and bounding box heads, the model employs a CLIP~\cite{clip} text embedding and uses an intermediate vision-language network for fusing both modalities. These advancements allow prompting the model with new classes at inference time, letting it detect objects it was not explicitly trained to detect.

In the realm of aerial view person detection, several methods such as Mobilenetv2~\cite{geraldes2019}, Faster R-CNN~\cite{hung2020faster}, SSPNet~\cite{hong2021sspnet}, and YOLOv3~\cite{speth2022deep} have been employed. These models are typically fine-tuned with bespoke datasets captured using drones and manually annotated. However, the significant effort required for recording and annotating such datasets prompts us to explore alternative approaches.


\section{EVALUATION}
\label{evaluation}
\subsection{DATASET}
\label{okutama}
For our evaluation, we consider the Okutama-Action dataset~\cite{2017okutamaaction} because it presents a challenging scenario and covers both our areas of interest: person detection and action classification.
In the Okutama-Action dataset, multiple persons are captured from aerial views using two drones at various times of the day. The dataset comprises different flights, each averaging about one minute in duration. Each camera frame depicts up to 10 individuals engaged in 12 distinct actions. These actions include running, walking, lying, sitting, standing, reading, drinking, pushing/pulling, carrying, calling, handshaking, and hugging.

\begin{figure}
    \centering
    \includegraphics[width=\linewidth]{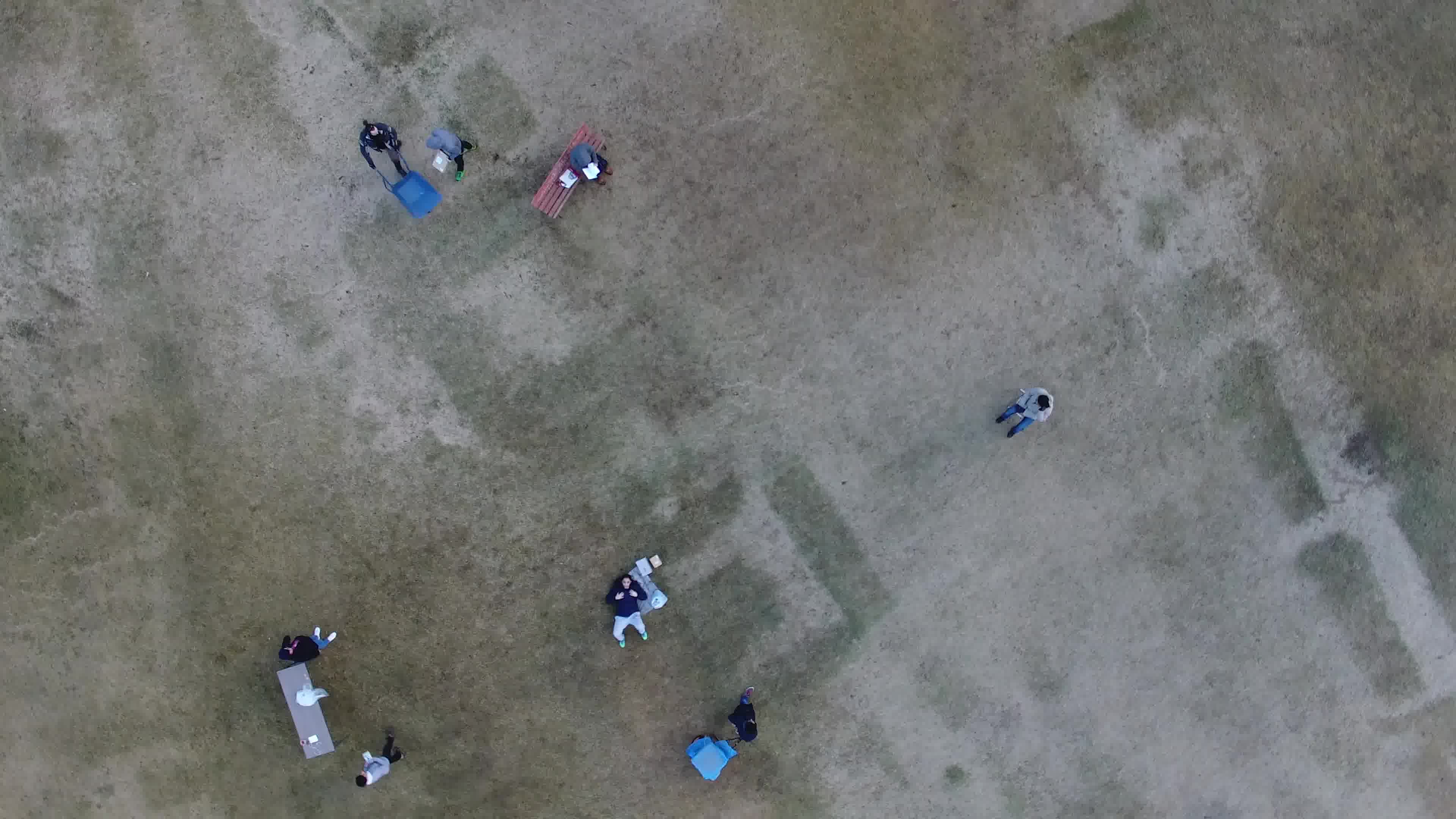}
    \caption{Sample image of the Okutama-Action dataset. The image shows eight persons, performing the actions calling, carrying, push/pulling, lying, reading, sitting, and standing.}
    \label{fig:sampleimg}
    \vspace{-5mm}
\end{figure}

While the dataset includes a total of 43 minutes of recordings, we're focusing solely on the test dataset, which consists of 10 flights totaling 9 minutes and 32 seconds. An example image from the dataset is illustrated in \Cref{fig:sampleimg}.

\subsection{PERSON DETECTION}
\label{persondetection}
\paragraph{GPT-4V Experiments}

In preliminary experiments, we wanted to evaluate whether GPT-4V can describe a drone image accurately. We prompted it with the image from \Cref{fig:sampleimg} and the following text prompt:

\begin{itemize}
\item  \textbf{Prompt:} ``You are an expert in classifying persons and their actions in pixelated images recorded from drones. Describe this picture in a few sentences. How many people are there, where are they in the image and what are they doing?"
\item \textbf{Response:} ``In the image, there appear to be five individuals and various objects that could be tables or benches. The people are dispersed across the frame with no two individuals close together, suggesting they may be engaging in independent activities. One person is sitting on a bench, another is near a table, and the others are standing. They seem to be in a public space, possibly a park, given the open grassy area and the park-like furniture visible. It's difficult to ascertain their specific actions due to the image's perspective and resolution, but they do not appear to be interacting with each other and are rather involved in separate tasks or leisure activities."
\end{itemize}

As the response suggests, GPT-4V can grasp the overall scene depicted in the image and has a basic understanding of what the people are doing and what might be the background. However, it struggles with accurately counting the number of people and identifying their locations within the image. Additional preliminary experiments also suggest that GPT-4V lacks the ability to determine the boundaries of individuals accurately. In most cases, the boundaries were significantly misaligned and individuals were not captured by them at all. We attempted various prompts, including requesting relative image coordinates for axis-aligned bounding boxes, center coordinates of individuals, and indexing an imaginary grid, as well as overlaying an actual grid onto the image. However, none of those methods resulted in a significant improvement in detection quality.

It is also worth mentioning that GPT-4V often refuses to give an answer because of safety policies~\cite{wu2024jailbreaking}. However, we could circumvent this issue by ``distracting'' it to reply in a specific JSON format. We assume that then, it interpreted the request more like a programming task, rather than a ``detect and classify humans'' task. However, it is likely that these safety-related fine-tuning steps are also negatively influencing the accuracy of GPT-4V.

\paragraph{YOLO-World Experiments}

To compensate for the person detection shortcomings of GPT-4V, we apply YOLO-World with pre-trained weights on the dataset. We prompt the model just with a single-word text-prompt `Person' and load the published weights YOLO-Worldv2-L-1280. We applied a basic non-maximum suppression to the predicted bounding boxes. Other than that, the predicted bounding boxes were calculated separately for each frame and we haven't applied any filtering, filling, smoothing or any other post-processing on them. 

\begin{figure}
    \centering
    \includegraphics[width=1\linewidth]{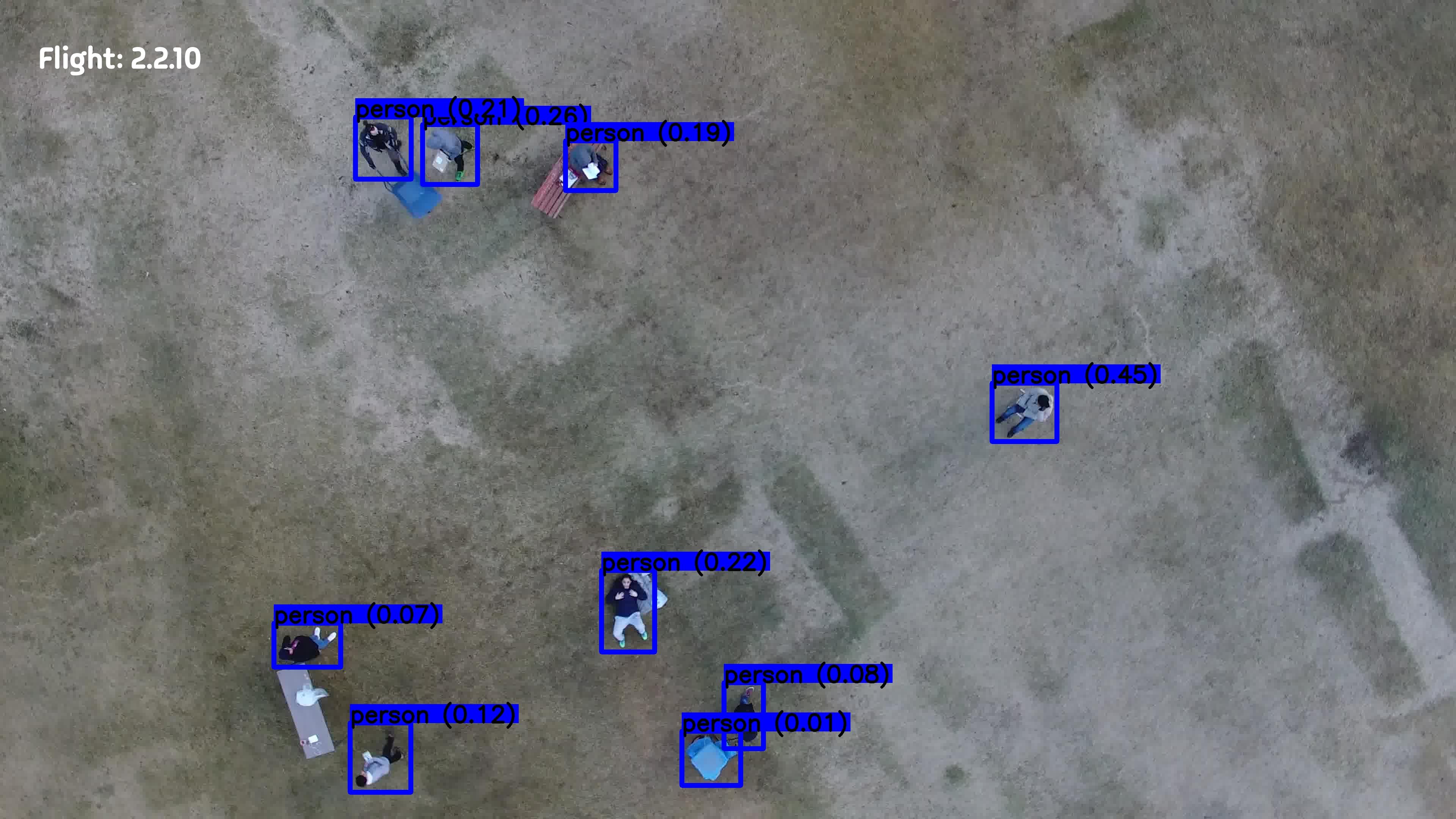}
    \caption{YOLO-World detection of aerial image. We prompt the model with a single class `Person' and loaded pre-trained weights. We uploaded a detection video of the full test dataset here: \url{https://www.youtube.com/watch?v=QntgkMKVuVQ}.}
    \label{fig:yoloworlddet}
    \vspace{-5mm}
\end{figure}

As depicted in \Cref{fig:yoloworlddet}, YOLO-World accurately detects the persons in the images. However, the confidence scores associated with the bounding boxes are relatively low, likely because the model was primarily trained with close-ups or group portraits. To address this, we filter the bounding boxes using a very low confidence threshold of 0.01. This approach results in some false positive detections, which can be subsequently filtered out by GPT-4V. We delve deeper into this issue in \Cref{actionrecognition}.

\begin{table}
    \centering
    \caption{YOLO-World zero-shot person detection performance.}
    \label{tab:yoloperformance}

    \begin{tabular}{l|cccc}
\hline
\textbf{flight} & \textbf{precision} & \textbf{recall} & \textbf{F1} & \textbf{mIOU} \\
\hline

2.2.10 & 0.643 & 0.530 & 0.572 & 0.228 \\
1.2.1 & 0.946 & 0.575 & 0.707 & 0.494 \\
2.1.9 & 0.888 & 0.880 & 0.868 & 0.622 \\
1.2.3 & 0.866 & 0.748 & 0.785 & 0.468 \\
1.1.9 & 0.802 & 0.839 & 0.791 & 0.507 \\
2.2.3 & 0.866 & 0.814 & 0.820 & 0.651 \\
2.2.1 & 0.369 & 0.317 & 0.333 & 0.117 \\
2.1.8 & 0.067 & 0.066 & 0.066 & 0.033 \\
1.2.10 & 0.674 & 0.642 & 0.654 & 0.281 \\
1.1.8 & 0.728 & 0.818 & 0.742 & 0.539 \\

    \end{tabular}
    \vspace{-5mm}
\end{table}

\Cref{tab:yoloperformance} depicts the performance of YOLO-World. We calculate the precision, recall and F1 scores of the detections. We classify a detection to be true if the bounding box is overlapping with the ground-truth bounding box by at least 10\%. This low percentage is justified because the ground-truth bounding boxes are not very accurate. Only one in every 10 frames was manually annotated, while the rest were interpolated, and their quality per flight varies drastically --
we could confirm that flight 2.2.1 and 2.1.8 had a very poor annotation, which is especially reflected in the mean Intersection Over Union (mIOU) column of \Cref{tab:yoloperformance}. The accuracy of YOLO-World stays consistent across the flights, which is shown in the rendered video linked in the caption of ~\Cref{fig:yoloworlddet}.

It is noticeable that, performance-wise, YOLO-World cannot compete with a traditionally trained YOLOv3 approach~\cite{speth2022deep}, especially regarding mean Intersection over Union (mIOU). However, the achieved performance may be sufficient for drone use cases where detecting a person is more important than the prediction of an accurate image location.

\subsection{ACTION RECOGNITION}
\label{actionrecognition}

In this section, we explore whether GPT-4V can classify the actions of the persons detected in the previous subsection. Our experimental setup is as follows: we select 10 images from each flight of the test dataset, and we utilize YOLO-World for person detection. For each prediction, we calculate the Intersection over Union (IOU) with all the ground-truth bounding boxes and assign the label of the bounding box with the highest IOU. If there is no overlap, we label the predicted bounding box as `no\_person'.

We define two recognition problems. Firstly, we assess whether GPT-4V can effectively filter out region proposals that do not contain a person, which is essentially a binary classification problem of `person' or `no\_person'. Secondly, we investigate whether GPT-4V can accurately determine the correct action class for each person, representing a 13-class classification problem (12 action classes and the `no\_person' token). For each problem, we calculate the 0/1 accuracy and the F1 score.

For each problem, we have conducted four experiments:

\begin{itemize}
\item \textbf{Experiment XXX}: We used the simplest prompt: ``Is there a person in this image? If yes, what activity is he or she doing? Return one of the following answers and nothing else: [action classes and `no\_person' token]."
\item \textbf{Experiment EXX}: We applied an `expert priming' by prepending the sentence: ``You are an expert in classifying persons and their actions in pixelated images recorded from drones.'' to the XXX prompt.
\item \textbf{Experiment EDX}: We requested an explanation for the recognized class by appending the following to the EXX prompt: ``Return the activity and a short statement why you think the person is doing this activity. Return in this format and nothing else: ["activity", "statement"]."
\item \textbf{Experiment EDS}: We incorporated the preceding and subsequent image in the time series to consider temporal information (each of them is 10 frames, or 0.3 seconds, apart of the image of interest). We hope that by adding the time dimension, actions like walking can be classified more accurately because the model would recognize the movement of the legs.
\end{itemize}


\begin{table}[htbp]
    \centering
    \caption{GPT-4V scores for person/non-person recognition and the recognition of 13 action classes.}
    \label{tab:scores}
    \begin{tabular}{l|cc|cc|cc|cc}
        \hline
        \textbf{exp} & \multicolumn{2}{c}{\textbf{valid responses}} & \multicolumn{2}{c}{\textbf{person rec}} & \multicolumn{2}{c}{\textbf{action rec}} \\
        & \textbf{valid/total} & \textbf{ratio} & \textbf{0/1} & \textbf{F1} & \textbf{0/1} & \textbf{F1} \\
        \hline
        
XXX & 436/447 & 0.975 & 0.725 & 0.362 & 0.248 & 0.255 \\
EXX & 435/447 & 0.973 & 0.724 & 0.381 & 0.266 & 0.274 \\

EDX & 436/447 & 0.975 & 0.734 & 0.333 & 0.262 & 0.268 \\
EDS & 360/447 & 0.805 & 0.750 & 0.483 & 0.277 & 0.344 \\

    \end{tabular}
\end{table}

The results of the four experiments are listed in \Cref{tab:scores}. In general, performance improves slightly with expert priming (Experiment EXX). Demanding an explanation from the model does not seem to enhance the recognition performance (Experiment EDX). Querying the model with an image sequence (Experiment EDS) delivered the best results but for this experiment 20\% of the queries were rejected by OpenAI due to their safety policies (see \Cref{tab:scores}, column `valid responses'). 

\begin{figure}
    \centering
    \includegraphics[width=1\linewidth]{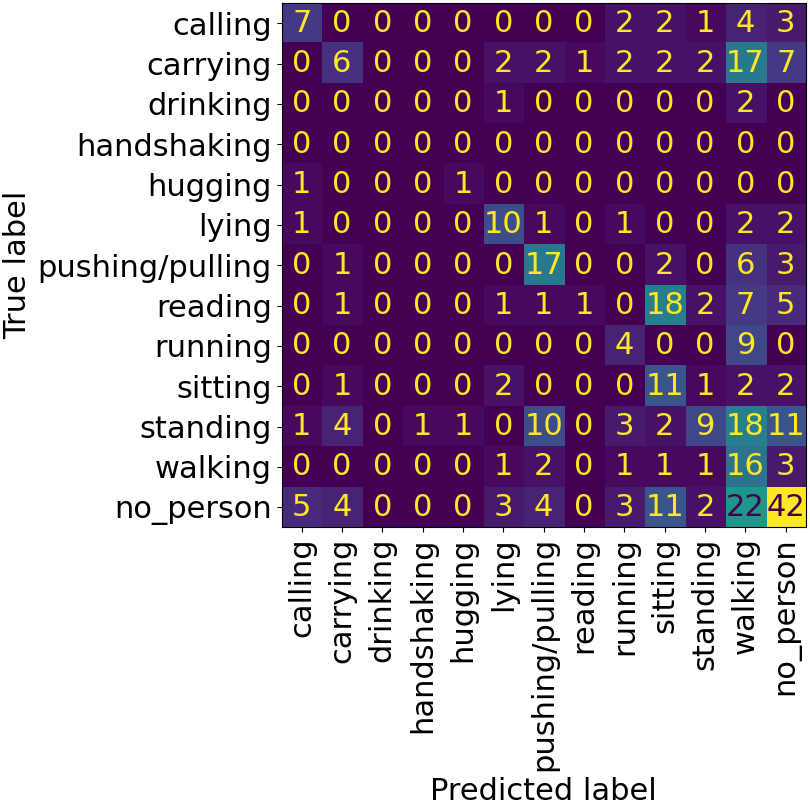}
    \caption{Confusion matrix for action recognition using GPT-4V.}
    \label{fig:confusionmatrix}
\end{figure}

\begin{figure}
    \centering
    \includegraphics[width=1\linewidth]{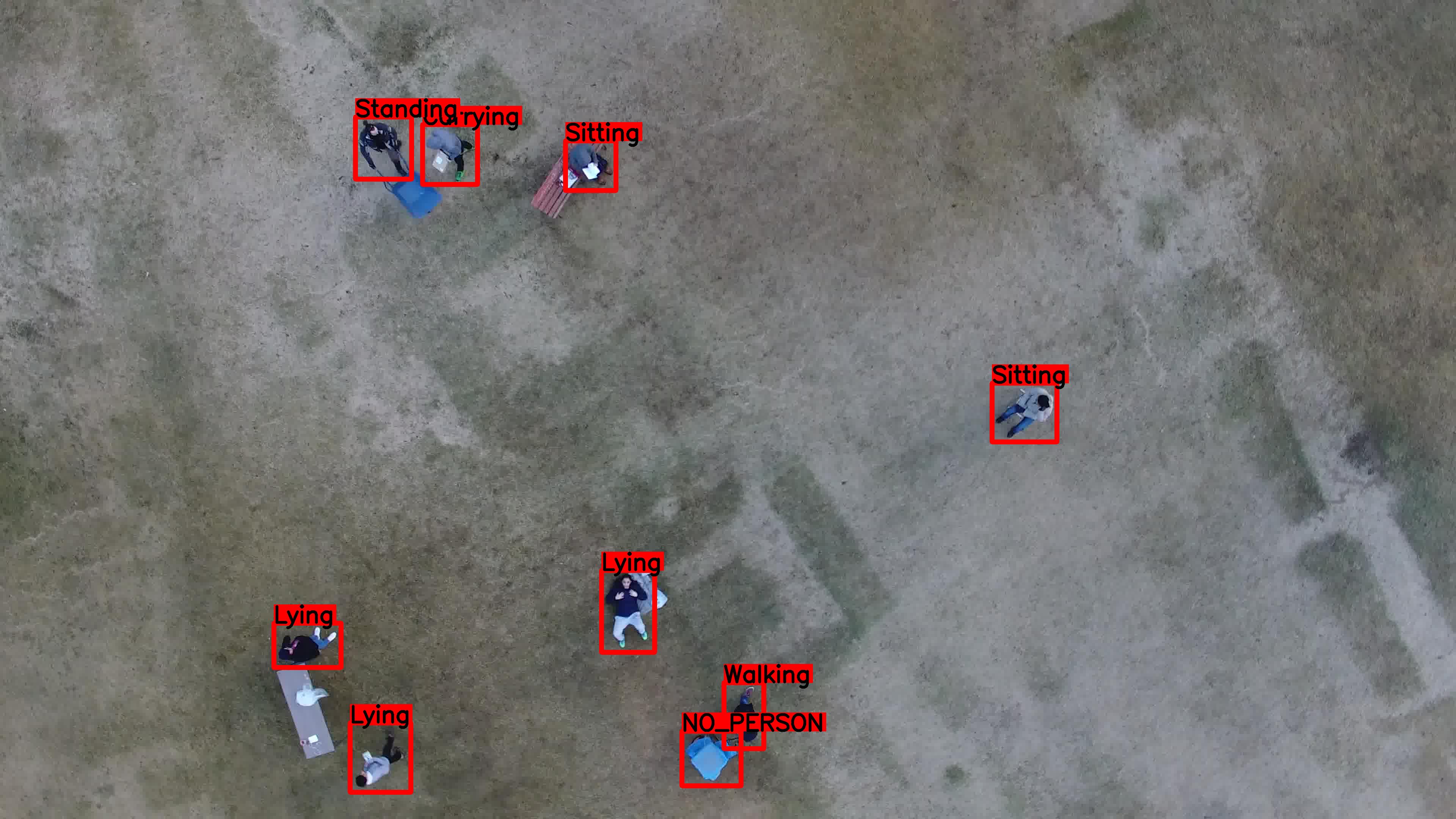}
    \caption{GPT-4V classification of YOLO-World region proposals.}
    \label{fig:gptbbclassification}
\end{figure}

However, the model can not predict action class accurately. This might be because of the inherently challenging nature of the dataset, where one person can perform multiple actions simultaneously, but only the most prominent action is labeled. For instance, a person could be carrying an item while walking. Predicting the class `walking' would be considered incorrect because the prominent class would be `carrying'. We have depicted a confusion matrix to highlight this characteristic of the dataset in \Cref{fig:confusionmatrix}. However, there are instances where unambiguous actions are also misclassified. For experiments EDX and EDS we verified the explanations the model returned for some examples. Sentences like ``The person appears to have one foot in front of the other, suggesting movement typical of walking." or ``The person appears to have one hand raised to their ear in a manner that suggests they are holding a phone, which indicates they might be making a call." indicates, that the model has a good understanding about the images and the prompted action classes. However, often the explanations were just wrong. We have illustrated the action recognition results for the previously discussed image in \Cref{fig:gptbbclassification}.

\section{CONCLUSIONS}
\label{conclusion}

We evaluated two zero-shot LMMs for their recognition applicability within a drone context. Our results suggest that YOLO-World achieved good detection performance. Subsequently, we employed GPT-4V to classify the predicted region proposals with action classes. However, the model struggles to provide accurate zero-shot predictions for the 12 action classes. Nevertheless, it could potentially be utilized to filter out unwanted region proposals or to provide a general description of the scenery.
While the accuracy may not yet be comparable to traditional approaches, there is a significant advantage in not having to train the models but simply prompting them. By changing just one word in the two prompts, a robot could be applicable for an entirely different use case, such as finding dogs or other objects of interest. 

In our future work, we aim to continue this work based on our findings to assess whether the two models are applicable in our rescue drone use case. Our preliminary investigation has revealed that GPT-4V possesses a detailed basic understanding of drone images, but it struggles to accurately determine the location and actions of people. The latter problem could potentially be addressed by prompting the model with some supervisory input in a few-shot learning manner, providing it with additional prior knowledge of the task.

\bibliographystyle{IEEEtran}
\bibliography{bib}

\end{document}